\documentclass[letterpaper]{article} 
\usepackage[preprint]{aaai2027}  
\usepackage[hyphens]{url}  
\usepackage{graphicx} 
\usepackage{natbib}  
\usepackage{caption} 
\usepackage{algorithm}
\usepackage{algorithmic}

\usepackage{newfloat}
\usepackage{listings}
\DeclareCaptionStyle{ruled}{labelfont=normalfont,labelsep=colon,strut=off} 
\floatstyle{ruled}
\newfloat{listing}{tb}{lst}{}
\floatname{listing}{Listing}

\usepackage{booktabs}
\usepackage{amsmath}
\usepackage{amssymb}

\title{\textsc{SCOPE}: Supply-Chain Operations through Coupled Policies for End-to-End Coordination}
\author{
    \textbf{Yunhao Liang}$^1$ \quad \textbf{Xianqi Cao}$^1$ \quad \textbf{Pujun Zhang}$^{*,1,2}$\\[1mm]
    \textbf{Yuan Qu}$^{*,1,2}$ \quad 
      \textbf{Yongzhi Qi}$^3$ \quad \textbf{Ningxuan Kang}$^3$ \quad \textbf{Max Z.J. Shen}$^{1,2}$
}
\affiliations{
    $^1$The University of Hong Kong, Hong Kong SAR, China\\[1mm]
    $^2$OptiMax AI, Hong Kong SAR, China\\[1mm]
    $^3$JD.com, China
}

\begin{document}

\maketitle

\begin{abstract}



Can supply-chain AI move beyond isolated decision modules toward unified operational planning? A complete replenishment plan specifies which products each location carries, which upstream facility supplies it, how often it is replenished, and how deliveries are routed. These decisions are operationally coupled: the selected assortment changes the demand and load passed to later stages; source assignment and replenishment frequency reshape the delivery requests; and route feasibility and cost, in turn, determine the system value of the earlier choices. Yet in modern supply chains, these decisions are often handled by separate departments and optimized through separate systems, which can lead to stockouts, inventory exposure, and avoidable transportation. We propose \textsc{SCOPE}: Supply-Chain Operations through Coupled Policies for End-to-End Coordination, a composite policy model that represents supply-chain entities as tokens, contextualizes them through a shared operational representation, and maps each token type to the corresponding decision interface. Each decision builds on the partial plan formed by earlier decisions while the completed plan is evaluated using a shared system-level utility. We instantiate this framework in urban fresh-retail replenishment, where service frequency, assortment, capacity pressure, and road-network routing interact strongly, and evaluate it on real operational data from Dingdong and JD.com, two large-scale supply chains operating at different replenishment echelons. Across both settings, \textsc{SCOPE} consistently outperforms methods that optimize each decision stage separately, as well as practice-oriented baselines commonly used in supply-chain operations. These results show that learning and coordinating cross-department operational couplings lead to more effective end-to-end supply-chain decisions.


\end{abstract}


\section{Introduction}
\label{sec:intro}

Supply chains are the operating systems of modern society: they coordinate products, capacity, inventory, and service commitments across organizations, facilities, and regions \citep{mentzer2001defining}, and their failures threaten economic security, public health, and the availability of essential goods \citep{baumgartner2020reimagining,whitehouse2021supplychains,christopher2004building,tang2006perspectives}. Yet the computational tools that operate these systems remain dominated by an older paradigm: each department solves its own optimization program, a demand forecaster here, a route solver there, while the system-level consequences of their interacting decisions are left unmodeled. Recent literature has therefore called for integrating machine learning into supply-chain operations \citep{baryannis2019predicting,brintrup2020supply}. We argue that the central AI challenge is not to forecast demand or optimize a downstream route more accurately, but to learn the couplings through which departmental decisions
determine each other's outcomes and coordinate them under system-level objective.

\begin{figure}[t]
    \centering
    \makebox[\columnwidth][c]{%
        \scalebox{1}[1.28]{%
            \includegraphics[
                width=1.10\columnwidth
            ]{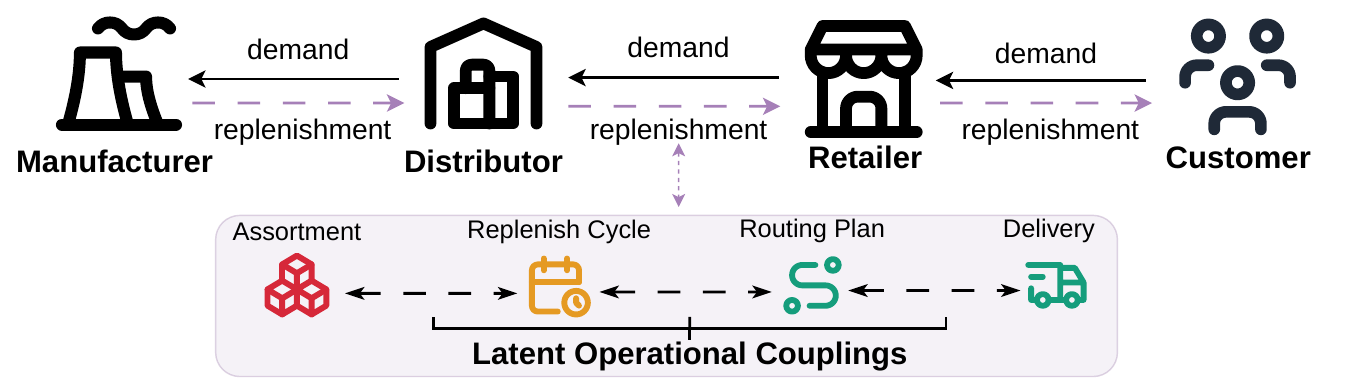}
        }%
    }
    \caption{Latent operational coupling in a replenishment supply chain. Assortment and replenishment-cycle decisions shape the demand released to logistics and hence the routing plan and delivery outcome; downstream execution costs, in turn, determine the system value of upstream choices.}
    \label{fig:latent-coupling}
\end{figure}


In the real modern supply chain, platforms are organized into functional decision layers: merchandising teams decide what products to carry, replenishment teams decide when and how often inventory moves, and logistics teams decide how vehicles and routes serve the resulting demand. This decomposition is organizationally natural but algorithmically incomplete, echoing a long literature on decentralized supply chains and incentive misalignment \citep{cachon2003coordination,cachon1999competitive,lariviere2001selling}: decisions that are locally optimal for one layer routinely shift cost, risk, or capacity burden onto another. 

Existing tools mirror this decomposition: assortment models, partial inventory--routing programs, and neural route solvers optimize local decision surfaces under fixed or
hand-specified cross-stage relationships\citep{kok2015assortment,coelho2014thirty,shaabani2022literature}. What they lack is a system view of the decision pipeline: existing methods optimize departments in isolation, ignoring upstream effects on downstream problems and whole system value. We call this missing cross-stage dependency \emph{latent operational coupling} (Figure~\ref{fig:latent-coupling}). For example, a high-demand product may become unattractive once its capacity consumption and routing burden are priced
in, while a longer replenishment cycle may reduce transportation cost but increase inventory exposure and lost-sales risk; similar tradeoffs arise throughout the pipeline. 
Because they are not directly observed, locally reasonable decisions can silently
transfer cost and risk across the system \citep{forrester1961industrial,sterman1989modeling,
lee1997information}. Existing methods either optimize within a fixed downstream instance or hand-specify fragments of the coupling; they do not learn and coordinate it across the whole supply-chain pipeline.

\textbf{The present work.} We formulate retail replenishment as a decision-learning problem with latent operational coupling. Assortment, source assignment, and replenishment interval decisions determine which demand is served, where it originates, and when it is released. Through known operational rules, these decisions induce the routing instances and feasible route sets that logistics must handle. We formalize an upstream plan through its induced system value, namely the best complete-pipeline utility attainable after completing its
remaining decisions.


To our best knowledge, \textsc{SCOPE} (Supply-Chain Operations through Coupled Policies for End-to-End Coordination) is the first learned decision framework to jointly reason over supply-chain entities and produce coordinated end-to-end operational plans under a shared system-level objective. It is designed around two objectives: (1) learning the latent operational couplings through which upstream choices reshape downstream problems and attainable system value, and (2) translating these couplings into complete, inspectable operational decisions.\textsc{SCOPE} encodes products, facilities, demand, geography, capacity, and cost context in a shared operational representation, while each policy interface
extends the partial plan formed by preceding decisions. This design is related to multi-task representation learning \citep{caruana1997multitask,ruder2017overview,
bommasani2021opportunities}, but targets end-to-end operational planning.



We validate \textsc{SCOPE} on two operational regimes chosen to place the same couplings at different structural positions. The first uses FreshRetailNet-50K, a public fresh-retail dataset ~\citep{wang2025freshretailnet}. The second, through collaboration with JD.com, uses a nationwide warehouse network in which selected demand must first be assigned from regional distribution centers (RDCs) to front distribution centers (FDCs). We train separate dataset-specific checkpoints of the same architecture: the experiments test whether the backbone-and-interface design absorbs a topological change through a single added interface, so the claim is structural reuse. Every method is evaluated as a complete pipeline against decomposed baselines with rule-based upstream decisions and strong classical, industrial, metaheuristic, and neural route solvers, together
with matched ablations and proxy stress tests. In summary, our contributions are:

\begin{enumerate}

\item \textbf{Problem:} a new decision-learning formulation of retail replenishment, in which upstream decisions induce downstream routing problems and complete plans are evaluated under a shared system-level utility;

\item \textbf{Model:} we introduce \textsc{SCOPE} model, which represents supply-chain entities as tokens, contextualizes them through a shared operational representation, and maps each token type to a corresponding decision interface for end-to-end coordination;

\item \textbf{Concept:} latent operational coupling, a modeling language for the implicit cross-stage dependencies among decisions over supply-chain entities;


\item \textbf{Results:} on real Dingdong and JD.com data, \textsc{SCOPE} outperforms strong decomposed pipelines at both warehouse and store levels and remains robust across broad proxy settings.

\end{enumerate}

\textbf{Social impact.} This work is a collaboration with JD.com, one of the world's largest e-commerce companies, whose nationwide fulfillment network motivated the problem: departmental optimizers for selection, replenishment, and logistics interact daily, yet no learned representation connects their decisions. Together we identified two objectives grounded in real operational needs. Learning operational couplings improves cross-department visibility, letting planners see how assortment and service-frequency choices propagate into capacity pressure, inventory exposure, and routing burden before costs materialize. Complete-pipeline decision learning supports joint improvement of replenishment frequency, warehouse assignment, and route execution. The FreshRetailNet experiments are fully reproducible from public data, and the JD evaluation preserves an auditable protocol despite data-sharing constraints, so that research on coupled supply-chain decision learning can continue beyond this collaboration.


\begin{figure*}[t]
    \centering
    \scalebox{1}[0.90]{%
        \includegraphics[
            width=0.98\textwidth
        ]{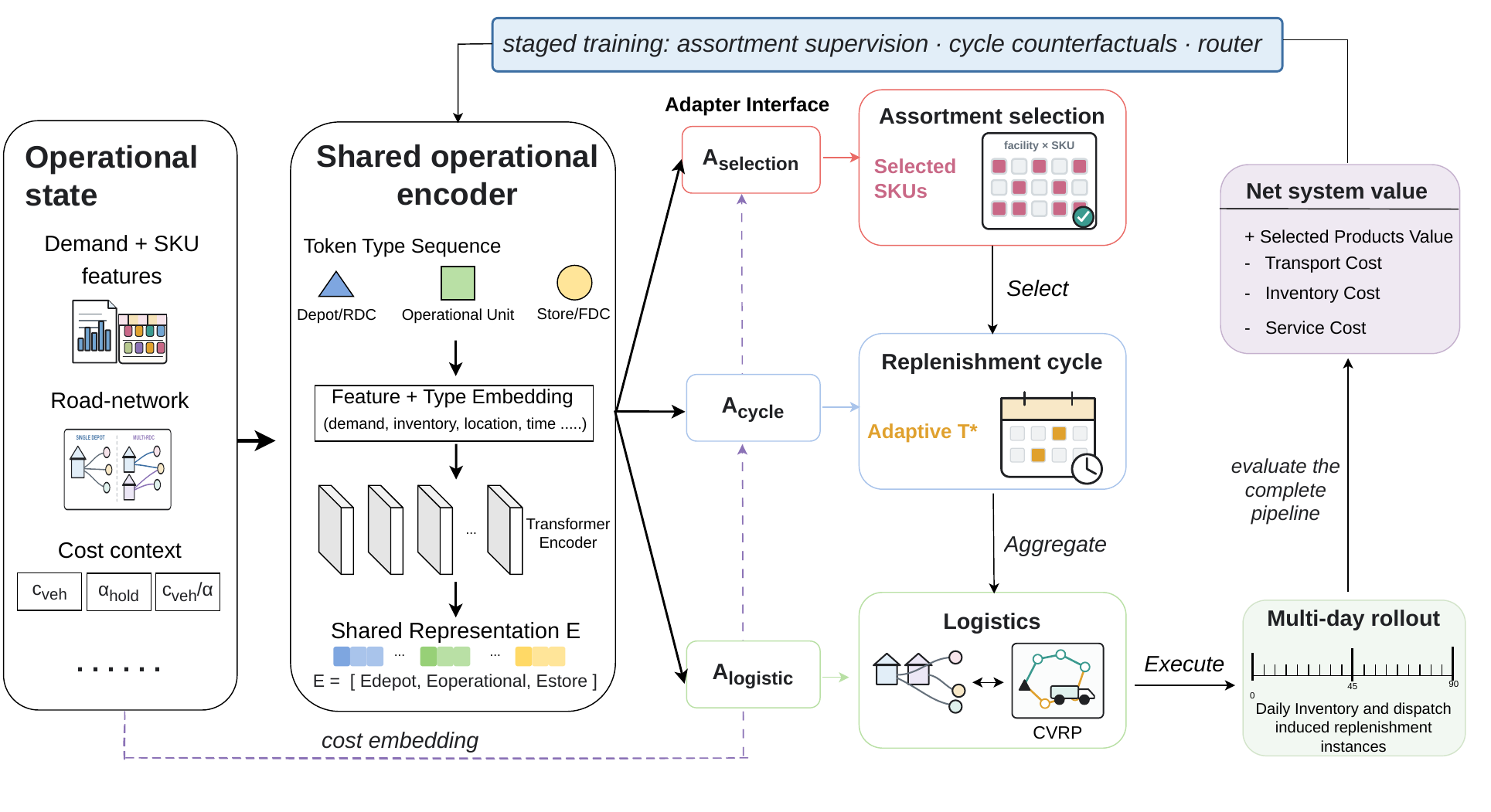}
    }
    \caption{
    \textsc{SCOPE} as a coupled supply-chain policy-optimization framework. A shared operational decision backbone supports department-facing policy interfaces for assortment selection, replenishment-cycle planning, and routing under a unified operational objective.
    }
    \label{fig:system-mechanism}
\end{figure*}

\section{Related Work}
\label{sec:relectedWork}

\textbf{Route-centric vehicle-routing solvers.}
Vehicle routing spans classical formulations, constructive heuristics, and rich VRP variants
\citep{dantzig1959truck,clarke1964scheduling,laporte1992vehicle,toth2014vehicle,vidal2013heuristics}. Representative route optimizers include nearest-neighbor construction \citep{rosenkrantz1977analysis}, industrial toolkits such as OR-Tools
\citep{ortools_routing}, and metaheuristics such as Hybrid Genetic Search
\citep{vidal2022hybrid}. Recent neural methods learn route-construction policies through attention, multi-start reinforcement learning, mixture-of-experts, or unified node--edge representations\citep{kool2019attention,kwon2020pomo,zhou2024mvmoe,meng2025uniteformer}.
We include representative methods from these families as route components in our baseline pipelines. We include representative methods from these families in our baseline pipelines.


\textbf{Machine learning for supply-chain operations.}
Recent surveys document growing applications of machine learning in supply-chain management \citep{baryannis2019predicting,younis2022applications}.
Existing work mainly addresses prediction and visibility, including supply-chain risk prediction\citep{baryannis2019predicting}, supplier-disruption prediction\citep{brintrup2020supply}, and graph learning for hidden-link prediction and planning benchmarks\citep{aziz2021data,kosasih2022machine,wasi2024supplygraph}.
A separate line learns policies for individual operational layers, including lost-sales, dual-sourcing, multi-echelon, and beer-game inventory problems \citep{gijsbrechts2022can,oroojlooyjadid2022beer,boute2022deep}, as well as end-to-end replenishment\citep{qi2023practical}. \citet{chang2025learning} learn hidden production functions from transaction-level supply-chain graphs. These methods, however, do not jointly coordinate decisions across supply-chain departments.

\textbf{Integrated supply-chain decision models.}
Operations research has long recognized that supply-chain decisions are coupled, including assortment planning \citep{kok2015assortment}, joint location--inventory models \citep{shen2003joint}, inventory-routing \citep{coelho2014thirty}, perishable inventory-routing
\citep{shaabani2022literature}, and integrated assortment--inventory or assortment--shelf-space--supplier models \citep{martinez2022integrated,sajadi2022integrated}. \citet{shen2007integrated} surveys the broader integrated supply-chain design literature, including models in which facility-location decisions account jointly for inventory and distribution costs.
Supply-chain coordination further studies how decentralized decisions can be aligned through contracts or related mechanisms \citep{cachon2003coordination}. These works establish the value of coordination, but typically hand-specify the coupling, cover only subsets of the operational pipeline, and rely on business quantities unavailable in many operational datasets.

\textbf{Positioning.}
Unlike route-centric solvers, we treat routing instances as outputs of learned upstream decisions and evaluate the complete supply-chain pipeline. Unlike prior supply-chain machine learning, we learn latent operational couplings and place them in joint policy optimization across departments. Unlike integrated optimization models, we learn a shared operational representation from demand, product, geography, and cost context, while handling unavailable business quantities through explicit proxies. \textsc{SCOPE} therefore formulates supply-chain operations as coupled decision learning, where multiple decision interfaces are built on a shared representation and optimized toward a unified system-level objective.

\section{\textsc{SCOPE} Framework}
\label{sec:framework}

 \subsection{Problem Formulation}
\label{sec:problem-formulation}

Retail replenishment comprises four dependent planning stages.  A stage denotes
decision dependency, not elapsed time.  Stage 1 selects the assortment, which
determines the demand and load considered when choosing a supply source.
Stage 2 assigns supply origins, which changes the source-dependent inputs to
interval planning and routing.  Stage 3 chooses replenishment intervals, where
an interval is the number of days between successive planned replenishments.
This decision determines service days and visit loads.  Stage 4 constructs
routes for the resulting requests.  Each nonterminal decision therefore changes
the problem faced by later stages and the utility attainable by the complete
plan.

Downstream locations are indexed by $i$, upstream supply nodes by $j$, and SKUs
by $p$.  A scenario $s$ contains facilities, physical road distances, demand
$d_{ip}$, SKU attributes, eligible supply links, vehicle limits, and
replenishment units.  The cost context $c$ specifies the dispatch, travel, and
inventory exposure tradeoffs known at planning time.

For each location $i$, assortment $A_i$ contains exactly $K$ SKUs, and $A$
collects all local assortments.  Source plan $\sigma$ assigns each location to
one eligible upstream node: $\sigma_i\in\mathcal J_i$, where $\sigma_i=j$ means
that $j$ supplies $i$.  Locations are partitioned into fixed replenishment units
$\mathcal M_z$.  Requests within a unit are released on the same service days.
Interval $T_z$ controls their frequency and accumulated demand.  We use $T$ for
the complete interval plan and $\mathcal T(s)$ for its feasible set.

Let $\mathbf w_p$ collect the available weight and volume load factors or
proxies for SKU $p$.  The selected daily load and nominal load for one
replenishment visit are
\begin{equation}
\begin{aligned}
  \bar{\mathbf q}_i(A)
    &=\sum_{p\in A_i}d_{ip}\mathbf w_p,\\
  \mathbf q_i(A,T)&=T_z\bar{\mathbf q}_i(A),
    \qquad i\in\mathcal M_z.
\end{aligned}
  \label{eq:selected-interval-load}
\end{equation}
A short interval creates frequent deliveries; a long interval creates larger
loads and greater inventory exposure.  These quantities are planning proxies
and may differ from realized freight.

Given $(A,\sigma,T)$, let $\mathcal R(s;A,\sigma,T)$ be the feasible set of
complete route plans over the planning horizon.  It enforces the selected
requests, assigned origins, service days, physical road distances, and vehicle
weight and volume limits.  Thus, $A$ determines what demand moves, $\sigma$
determines where it originates, and $T$ determines when and at what visit load
it is delivered.  These decisions jointly determine the routing problems and
their feasible routes.

Let $\mathcal Y(s)$ contain exactly the complete plans
$y=(A,\sigma,T,R)$ satisfying $|A_i|=K$ and
$\sigma_i\in\mathcal J_i$ for every $i$, $T\in\mathcal T(s)$, and
$R\in\mathcal R(s;A,\sigma,T)$.  We evaluate $y$ by the daily proxy utility
\begin{equation}
  U(s,c;y)=V(A)-L_{\mathrm{cov}}(A)
    -C_{\mathrm{rt}}(R;c)-C_{\mathrm{inv}}(A,T;c),
  \label{eq:proxy-utility}
\end{equation}
where $V$ measures assortment value, $L_{\mathrm{cov}}$ penalizes expected
demand outside the selected assortment, $C_{\mathrm{rt}}$ is transportation
cost, and $C_{\mathrm{inv}}$ measures interval-induced inventory exposure.  All
horizon totals are divided by the horizon length before entering $U$.
$L_{\mathrm{cov}}$ is an uncovered-demand proxy, distinct from realized lost
sales in a dynamic inventory simulation.

For each decision prefix below, $\mathcal Y(s\mid\cdot)$ contains the complete
plans in $\mathcal Y(s)$ that preserve the decisions already fixed.  We use the
same \emph{operational completion value} at all three boundaries:
\begin{equation}
\begin{aligned}
  Q_{s,c}(A)
    &:=\max_{y\in\mathcal Y(s\mid A)}U(s,c;y),\\
  Q_{s,c}(A,\sigma)
    &:=\max_{y\in\mathcal Y(s\mid A,\sigma)}U(s,c;y),\\
  Q_{s,c}(A,\sigma,T)
    &:=\max_{y\in\mathcal Y(s\mid A,\sigma,T)}U(s,c;y).
\end{aligned}
  \label{eq:completion-value}
\end{equation}
The three lines evaluate one value concept.  The
remaining decisions are $(\sigma,T,R)$, $(T,R)$, and $R$, respectively.  Thus,
each earlier decision changes both the remaining completion problem and the
complete-plan value attainable from it.  We call this dependence
\emph{operational coupling}.  Assortment changes the demand and load passed to
later decisions; source assignment changes origins and source-dependent
inputs; and interval planning changes service days, visit loads, and feasible
routes.  Alternative completion problems and their optimized values are not
recorded in operational data and are costly to compute.

For one scenario, the operational problem is
\begin{equation}
  y^*(s,c)\in\arg\max_{y\in\mathcal Y(s)} U(s,c;y),
  \label{eq:scenario-problem}
\end{equation}
whose optimal value is
$\max_{A:\,|A_i|=K\ \forall i}Q_{s,c}(A)$.  Repeating this search as conditions
change motivates a conditional policy $\Pi_\theta$:
\begin{equation}
\begin{aligned}
  \theta^*\in\arg\max_\theta\quad
    &\mathbb E_{(s,c)\sim\mathbb P}
      \left[U\!\left(s,c;\Pi_\theta(s,c)\right)\right]\\
  \mathrm{s.t.}\quad &\Pi_\theta(s,c)\in\mathcal Y(s).
\end{aligned}
  \label{eq:policy-learning}
\end{equation}
Equation~\eqref{eq:policy-learning} defines the policy-level target.  SCOPE
approximates this target with staged surrogates and selects among the resulting
composite policies using validation pipeline utility.

\subsection{Model Overview}
\label{sec:model-overview}

SCOPE operationalizes this coupling with a conditional composite policy.  Each
stage extends the current decision configuration: $A$ conditions source
assignment,
interval planning, and routing; $\sigma$ further conditions interval planning
and routing; and $T$ completes the input to routing.  SCOPE then groups the
selected requests by assigned source and service day, and a neural routing
policy produces $R$.  The resulting complete plan is evaluated by the common
utility $U$.  Figure~1 links this flow to Eq.~\eqref{eq:scenario-problem}.

\noindent\textbf{Scenario representation and sequential policies.}
A shared operational backbone $B_\theta$ maps facilities and physical road
distances from $s$ to $\mathbf E=B_\theta(s)$.  This embedding set contains
$\mathbf h_i$ for downstream locations and $\mathbf h_j$ for upstream nodes.
A second encoder maps $c$ to $\mathbf z_c=e_\theta(c)$.  The shared
representation supplies scenario context; cross-stage dependence enters
through the explicit conditioning on earlier decisions.  We use $\theta$ for
all trainable parameters.

Superscripts $A$, $\sigma$, $T$, and $R$ label an assortment scorer, source
scorer, interval classifier, and autoregressive routing decoder.  The complete
policy follows the same order as the formulation:
\begin{equation}
\begin{aligned}
  A      &=\pi^A_\theta(s,\mathbf E,\mathbf z_c),\\
  \sigma &=\pi^\sigma_\theta
             (s,\mathbf E,\mathbf z_c,A),\\
  T      &=\pi^T_\theta
             (s,\mathbf E,\mathbf z_c,A,\sigma),\\
  R      &=\pi^R_\theta(s,\mathbf E,A,\sigma,T).
\end{aligned}
  \label{eq:scope-factorization}
\end{equation}
These steps return $\Pi_\theta(s,c)=(A,\sigma,T,R)$.  SCOPE does not calculate
the ideal completion value in Eq.~\eqref{eq:completion-value}.  Instead,
stage-specific surrogates train the conditional policies, and validation
utility selects their composition.  The same construction applies to other
staged systems in which each earlier decision reshapes the remaining problem.

The assortment scorer combines $\mathbf h_i$, projected SKU attributes
$\mathbf u_p=P_\theta(x_p)$, cost context, and demand, value, and load features,
then selects the top $K$ SKUs.  For source assignment, the scorer compares
eligible nodes using facility embeddings, link features, and selected load.
Interval selection pools these features within $\mathcal M_z$ and chooses one
candidate $\tau\in\{1,\ldots,T_{\max}\}$.

\noindent\textbf{Routing problem construction and solution.}
Before decoding, SCOPE groups the selected requests by service day and assigned
origin.  Each resulting routing subproblem specifies its active locations,
loads, vehicle limits, and physical distance matrix.  City scenarios use fixed
spatial grouping; warehouse routing subproblems are partitioned by $\sigma$.
The decoder uses capacity masks and multi-start decoding, retaining the
lowest-cost route among its feasible rollouts under the implemented metric.
It therefore serves as a feasible approximate executor, and its realized
utility need not attain the maximum in Eq.~\eqref{eq:completion-value}.

\subsection{Model Training}
\label{sec:model-training}

SCOPE approximates Eq.~\eqref{eq:policy-learning} with policy-specific
surrogates and validation pipeline utility.

\noindent\textbf{Upstream supervision.}
The assortment policy first learns from SKU targets that combine coverage,
value, load, and inventory exposure proxies.  It then evaluates one-for-one
swaps through the complete downstream pipeline and retains a swap only when
proxy utility improves while the required coverage and value are preserved.
These comparisons provide local preferences among nearby assortments.

The source policy considers only eligible nodes and uses the nearest feasible
node as a geometric warm-start target.  This promotes a physically plausible
assignment but does not identify the source plan with the highest operational
completion value.
For each training scenario and cost context, SCOPE also evaluates candidate
intervals with one interval shared by all active units.  The best globally
fixed interval provides a common warm-start label, while the deployed policy
predicts a separate interval for each unit.  This scan supplies no unit-specific
alternative labels.  Assortment replacements and global interval candidates
receive pipeline comparisons; source assignment receives
feasibility-constrained geometric supervision.

\noindent\textbf{Router specialization.}
For each day-specific routing subproblem $\rho$, OR-Tools is used offline to
produce a reference sequence $r^{\mathrm{ref}}_\rho$.  The route decoder is
trained by imitation:
\begin{equation}
  \mathcal L_R=-\sum_\rho\sum_m
    \log \pi^R_\theta
    (r^{\mathrm{ref}}_{\rho,m}\mid
     r^{\mathrm{ref}}_{\rho,<m},\mathbf E,A,\sigma,T).
  \label{eq:route-imitation}
\end{equation}
Each subproblem contains the source, service day, selected requests, loads, and
vehicle limits determined by the earlier decisions.  Route imitation supplies
no full-pipeline value target for alternative decision configurations, and the
resulting routes remain approximate.  At inference, SCOPE executes these
subproblems without OR-Tools.

\noindent\textbf{Full pipeline selection.}
The implementation keeps a separate loss for each policy.  We compare
candidate checkpoints $\mathcal C$ by running the complete policy on the
validation set:
\begin{equation}
  \widehat\theta\in\arg\max_{\theta\in\mathcal C}
  \frac{1}{|\mathcal D_{\mathrm{val}}|}
  \sum_{(s,c)\in\mathcal D_{\mathrm{val}}}
  U\!\left(s,c;\Pi_\theta(s,c)\right).
  \label{eq:pipeline-selection}
\end{equation}
Branch-specific surrogates fit candidate policies.  Validation utility $U$
then selects among the resulting composite checkpoints without supplying
gradients to the policy heads.  Proxy definitions, normalization, and
surrogate targets appear in the experimental protocol or appendix.  Thus,
SCOPE coordinates the conditional policies using complete-pipeline evaluations
without explicitly estimating $Q_{s,c}$ in Eq.~\eqref{eq:completion-value}.

\section{Experiments}
\label{sec:exp}

\begin{table*}[t]
\centering
\renewcommand{\arraystretch}{1.04}
\setlength{\tabcolsep}{3pt}
\begin{tabular*}{\textwidth}{@{\extracolsep{\fill}}lccccccccccc@{}}
\toprule
& & & \multicolumn{4}{c}{Dingdong (depot--store)} & & \multicolumn{4}{c}{JD.com (RDC--FDC)} \\
\cmidrule(lr){4-7}\cmidrule(lr){9-12}
Pipeline & Assort. & Cycle & $T$ & $U_{\mathrm{ref}}\uparrow$ & Gap$\downarrow$ & Cov.$\uparrow$ & & $T$ & $U_{\mathrm{ref}}\uparrow$ & Gap$\downarrow$ & Cov.$\uparrow$ \\
\midrule
\textsc{SCOPE} & $\checkmark$ & $\checkmark$ & $\checkmark$ & \textbf{206329.79} & -- & 0.948 & & $\checkmark$ & \textbf{293547.12} & -- & 0.427 \\
\midrule
\multicolumn{12}{@{}l}{\emph{TopDemand assortment with best global fixed cycle}} \\
TopDemand--OR-Tools & TopDemand-K & BestFix & 6 & 206276.14 & 53.65 & 0.948 & & 2 & 288734.46 & 4812.65 & 0.429 \\
TopDemand--NN & TopDemand-K & BestFix & 6 & 206271.86 & 57.93 & 0.948 & & 2 & 287021.11 & 6526.01 & 0.429 \\
TopDemand--HGS & TopDemand-K & BestFix & 7 & 206253.31 & 76.48 & 0.948 & & 2 & 288736.72 & 4810.39 & 0.429 \\
TopDemand--UniteFormer & TopDemand-K & BestFix & 7 & 206248.27 & 81.51 & 0.948 & & 3 & 287320.33 & 6226.78 & 0.429 \\
TopDemand--AM & TopDemand-K & BestFix & 7 & 206227.23 & 102.56 & 0.948 & & 4 & 282564.81 & 10982.30 & 0.429 \\
TopDemand--POMO & TopDemand-K & BestFix & 7 & 206197.59 & 132.20 & 0.948 & & 4 & 285398.48 & 8148.64 & 0.429 \\
TopDemand--MVMoE & TopDemand-K & BestFix & 7 & 206180.81 & 148.98 & 0.948 & & 3 & 284890.42 & 8656.70 & 0.429 \\
\midrule
\multicolumn{12}{@{}l}{\emph{Daily replenishment baselines}} \\
Daily--HGS & TopDemand-K & Daily & 1 & 205932.38 & 397.41 & 0.948 & & 1 & 283074.28 & 10472.83 & 0.429 \\
Daily--OR-Tools & TopDemand-K & Daily & 1 & 205931.78 & 398.01 & 0.948 & & 1 & 283074.28 & 10472.83 & 0.429 \\
Daily--NN & TopDemand-K & Daily & 1 & 205911.85 & 417.94 & 0.948 & & 1 & 280156.60 & 13390.51 & 0.429 \\
\midrule
\multicolumn{12}{@{}l}{\emph{Alternative assortment policies with best global fixed cycle}} \\
TopValue--OR-Tools & TopValue-K & BestFix & 6 & 205569.95 & 759.84 & 0.945 & & 2 & 288973.15 & 4573.96 & 0.427 \\
TopValue--HGS & TopValue-K & BestFix & 6 & 205556.16 & 773.63 & 0.945 & & 2 & 288973.15 & 4573.96 & 0.427 \\
Random--OR-Tools & Random-K & BestFix & 6 & 170016.63 & 36313.15 & 0.888 & & 5 & 65370.22 & 228176.90 & 0.103 \\
Random--HGS & Random-K & BestFix & 7 & 170003.01 & 36326.78 & 0.888 & & 5 & 65371.79 & 228175.32 & 0.103 \\
\bottomrule
\end{tabular*}
\caption{Complete-pipeline results on depot-to-store (Dingdong, 18 90-day scenarios) and RDC-to-FDC (JD.com, 18 91-day real-FDC subset scenarios). $U_{\mathrm{ref}}$ is larger-is-better; Gap$=U_{\mathrm{ref}}^{\textsc{SCOPE}}-U_{\mathrm{ref}}^{\mathrm{method}}$. BestFix is one global fixed cycle over $T=1,\ldots,7$, not a per-episode oracle; $T$ is the selected cycle on each dataset. A checkmark denotes a learned decision; Gap ``--'' means not applicable.}
\label{tab:main-results}
\label{tab:plan-a-main-results}
\label{tab:jd-main-results}
\end{table*}

\begin{table*}[t]
\centering
\renewcommand{\arraystretch}{1.06}
\setlength{\tabcolsep}{3pt}
\begin{tabular*}{\textwidth}{@{\extracolsep{\fill}}llccccccc@{}}
\toprule
& & \multicolumn{3}{c}{Dingdong (depot--store)} & & \multicolumn{3}{c}{JD.com (RDC--FDC)} \\
\cmidrule(lr){3-5}\cmidrule(lr){7-9}
Variant & Change & $U_{\mathrm{ref}}\uparrow$ & $\Delta$ & Cov.$\uparrow$ & & $U_{\mathrm{ref}}\uparrow$ & $\Delta$ & Cov.$\uparrow$ \\
\midrule
\textsc{SCOPE} & full staged pipeline & \textbf{206329.79} & 0.00 & 0.948 & & \textbf{293547.12} & 0.00 & 0.427 \\
w/o assortment interface & replace with TopDemand-K & 206275.89 & -53.90 & 0.948 & & 293261.53 & -285.59 & 0.429 \\
w/o service-cycle interface & fixed daily cycle ($T{=}1$) & 205903.42 & -426.36 & 0.948 & & 279026.38 & -14520.74 & 0.427 \\
w/o learned-router & remove reference-route warm-up & 206319.11 & -10.68 & 0.948 & & 292738.46 & -808.65 & 0.427 \\
w/o cycle supervision & remove counterfactual training & 206292.13 & -37.65 & 0.948 & & 288859.25 & -4687.86 & 0.427 \\
Direct joint training & train all modules from scratch & 173907.60 & -32422.18 & 0.894 & & 293472.61 & -74.50 & 0.427 \\
\bottomrule
\end{tabular*}
\caption{Decision ablations under the same protocols as Table~\ref{tab:main-results} (Dingdong: 18 90-day scenarios; JD.com: 18 91-day scenarios). $U_{\mathrm{ref}}$ is larger-is-better; $\Delta$ is the signed change relative to full \textsc{SCOPE}. Assignment remains part of the learned JD.com pipeline but is not treated as a separate ablation.}
\label{tab:ablation}
\label{tab:plan-a-ablation}
\label{tab:jd-ablation}
\end{table*}

Replenishment recurs at every echelon of a supply chain in the same abstract form, a shared decision grammar: an upstream supply node serves a set of downstream demand nodes, and the operator must decide what to serve, how often to replenish, and how to route the deliveries. This recurring structure motivates our evaluation along three dimensions. (i) \emph{Coupling}: do the latent operational couplings among departmental decisions matter, i.e., does learning these couplings and optimizing the complete assortment--cycle--routing pipeline jointly outperform decomposed pipelines that pair rule-based upstream decisions with strong route solvers? (ii) \emph{Generality}: does the same joint model remain effective across supply-chain scales and echelons? (iii) \emph{Mechanism and limits}: which learned interfaces drive the gains, and under which cost regimes do they fail? The main results address coupling and generality; the ablations and robustness studies isolate the underlying mechanisms and failure regimes.

\subsection{Data}
\label{sec:exp-data}
\begin{figure}[t]
\centering
\includegraphics[width=0.98\linewidth]{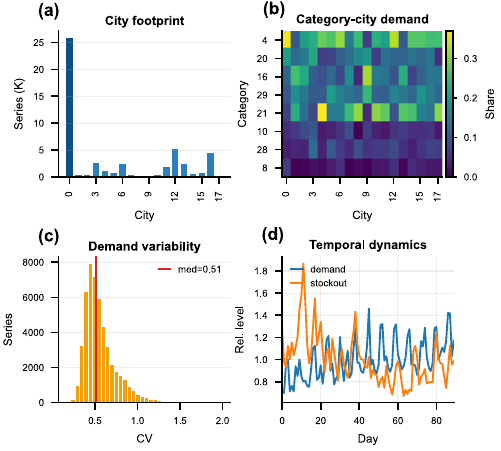}
\caption{FreshRetailNet-50K exhibits the heterogeneity that couples the three decisions: (a) skewed store-SKU footprint across 18 cities, (b) city-specific category mix, (c) long-tailed per-series demand volatility, (d) non-stationary demand with a censored stockout signal.}
\label{fig:freshretail-data-characteristics}
\end{figure}

We ground this evaluation in fresh retail, where assortment, replenishment frequency, inventory exposure, and distribution are tightly coupled~\citep{vandonselaar2006inventory,minner2010periodic,shaabani2022literature}, using two industrial level operational datasets at different supply-chain echelons. At the depot-to-store echelon, FreshRetailNet-50K~\citep{wang2025freshretailnet} is a rare public record of large-scale fresh-retail operations from Dingdong, spanning 898 stores, 18 cities, 865 SKUs, and 90 days. At the upper echelon of supply chain scenario, our JD.com data capture a nationwide, industrial multi-warehouse network with 8 RDCs, 43 FDCs, 1,110 SKUs, and 91 dated snapshots. Such facility-level inventory, demand, and network records are rarely available for academic evaluation; they expose warehouse-scale assortment and an assignment-dependent multi-depot topology, providing a stringent test of whether the same model generalizes beyond last-leg replenishment. Under the NDA, identities are masked and business quantities rescaled while the operational decision structure is preserved. Figure~\ref{fig:freshretail-data-characteristics} previews the public dataset's heterogeneity in store--SKU coverage, category mix, demand, and stockout signals. Full dataset details are provided in the Supplementary Material.

\subsection{Experimental Setup}
\label{sec:exp-setup}


Every method is evaluated as a complete assortment--cycle--routing pipeline over the full planning horizon, under one shared objective. We report mean reference-adjusted operational value $U_{\mathrm{ref}}$ (larger is better), together with transportation-plus-inventory cost, demand coverage, and vehicles per day. \textsc{SCOPE} is one learned joint pipeline; baselines are decomposed pipelines that cross three assortment rules (TopDemand-K, TopValue-K, Random-K), globally scanned fixed cycles (BestFix / Daily), and seven route executors ranging from nearest-neighbor and OR-Tools~\citep{ortools_routing} to recent neural routers~\citep{vidal2022hybrid,kool2019attention,kwon2020pomo,zhou2024mvmoe,meng2025uniteformer}. The top-$K$ assortment rules mirror JD.com's deployed prediction-then-ranking practice~\citep{shen2025jdcom}, so they are industrial baselines. Cost coefficients follow reported Chinese cold-chain vehicle costs and textbook carrying-cost conventions~\citep{feng2024optimization,pan2024cold,silver1998inventory}. Remaining details are in the Supplementary Material.

\subsection{Main Results Across Two Replenishment Echelons}
\label{sec:exp-city}
\label{sec:jd-validation}

Table~\ref{tab:main-results} reports the complete-pipeline comparison on both echelons. Across the Dingdong scenarios, \textsc{SCOPE} achieves the highest $U_{\mathrm{ref}}$ (206329.79), exceeding every decomposed combination of a top-ranked assortment rule, a globally optimized fixed cycle, and a strong classical or neural router. This is a demanding comparison: prediction-then-ranking assortment is used in JD.com's deployed fulfillment practice \citep{shen2025jdcom}, while OR-Tools, HGS, and the learned routers are competitive routing methods. Yet, with TopDemand-K and BestFix held fixed, replacing OR-Tools with any of the six alternative routers changes the gap only from $53.65$ to $148.98$; by comparison, imposing daily replenishment increases it to $398.01$, and changing the assortment policy increases it further. Thus, optimizing strong departmental policies separately is insufficient: once assortment and cycle have formed the routing instance, a stronger router cannot recover value lost through their interaction. \textsc{SCOPE}'s consistent lead provides direct evidence for the first question, that learning the latent operational couplings among assortment, replenishment frequency, and routing matters. The JD.com results then test whether this finding persists beyond the last leg. Despite moving to a larger RDC-to-FDC network with warehouse-scale SKU volumes and assignment-based instance formation, \textsc{SCOPE} again ranks first, reaching 293547.12 and outperforming the strongest decomposed pipeline by $4573.96$ (1.58\%) at essentially the same coverage. The same pattern is more pronounced there: daily replenishment loses $10472.83$, whereas changing the router within a fixed upstream policy cannot close the gap. Standing out on both a public depot-to-store benchmark and a proprietary industrial RDC-to-FDC network shows that the benefit is not tied to one dataset, scale, or replenishment echelon, answering the second question on generality. Full matrices and implementation details are provided in the Supplementary Material.

\subsection{Ablations}
\label{sec:exp-ablation}

Table~\ref{tab:ablation} reports matched decision ablations on both datasets, each replacing one learned interface or one training stage while keeping the rest of the pipeline fixed. The decision ablations reveal a consistent pattern across the two replenishment echelons. Removing the learned interfaces produces the largest performance degradation on both Dingdong ($-426.36$) and JD.com ($-14520.74$), while removing counterfactual cycle supervision causes a further substantial loss on JD.com ($-4687.86$). Replacing the learned assortment policy with TopDemand-K has a smaller but consistently negative effect. These results suggest that system value arises from learning the latent operational couplings among assortment, replenishment, and routing decisions. The ablations therefore support the central claim of Table~\ref{tab:main-results}: end-to-end performance depends on coordinating these interdependent decisions rather than optimizing each department in isolation. The training ablation highlights a practical difficulty in real supply-chain optimization. Because assortment, replenishment cycle, and routing jointly determine the final system outcome, a performance drop is difficult to attribute to any single decision module or to a poor combination of the three. On Dingdong, direct joint training from scratch converges to a substantially worse, lower-coverage policy, losing 32422.18 value units and reducing coverage from 0.948 to 0.894. On JD.com, however, direct joint training finishes only 74.50 behind staged \textsc{SCOPE}. Together, these results validate the training ablation by showing that the optimization strategy can materially affect the learned joint policy, although the magnitude of its effect varies across data regimes.

\subsection{Robustness and Failure Regimes}
\label{sec:exp_proxy}


To test whether the gains depend on proxy calibration, we
keep each checkpoint fixed and perturb the assortment-value, route-cost, holding-cost, and load scales. \textsc{SCOPE} ranks first in 30 of 35 value--cost settings and all seven load settings on Dingdong, and in all 35 value--cost settings and five of seven load settings on JD.com. Losses on Dingdong occur only at a $0.25\times$ assortment-value scale, where the objective approaches routing alone; those on JD.com occur when load perturbations change the induced capacity regime. Thus, the coupling advantage is robust across broad proxy ranges but can weaken when upstream value becomes negligible or the capacity regime shifts. Complete grids and numerical results are reported in the Supplementary Material.



\section{Conclusion}

This paper studies supply-chain operations as a coordinated decision-learning problem and introduces \textsc{SCOPE}, a composite policy model that represents supply-chain entities as typed tokens, contextualizes them through a shared operational representation, and realizes assortment, source assignment,
replenishment-cycle planning, and routing as sequentially coupled decision interfacese. \textsc{SCOPE} addresses two central objectives: learning the latent operational couplings through which upstream decisions reshape downstream problems, and using these couplings to produce complete, inspectable plans under a shared system-level utility---whereas prior methods typically optimize a fixed downstream instance or hand-specify only part of the coupling. To demonstrate these capabilities, we evaluated \textsc{SCOPE} on city-scale Dingdong replenishment and a nationwide JD.com warehouse network, where it consistently
outperformed decomposed operational pipelines across two replenishment echelons. More broadly, this work suggests a path toward supply-chain AI systems that improve whole-pipeline reliability.

\bibliography{references}


\end{document}